\pdfoutput=1

\documentclass[11pt]{article}

\usepackage[preprint]{acl}

\usepackage{times}
\usepackage{latexsym}
\usepackage{multirow}
\usepackage{booktabs}
\usepackage{algorithm}
\usepackage{amsmath}
\usepackage{algpseudocode}
\usepackage{xcolor}
\usepackage{amsmath}
\usepackage{amssymb}

\usepackage[utf8]{inputenc}
\usepackage{tcolorbox}
\usepackage{caption}

\tcbuselibrary{skins}          

\usepackage[T1]{fontenc}

\usepackage[utf8]{inputenc}

\usepackage{microtype}

\usepackage{inconsolata}

\usepackage{graphicx}

\title{DeRAGEC: Denoising Named Entity Candidates \\ with Synthetic Rationale for ASR Error Correction}

\author{Solee Im$^*$$^1$, Wonjun Lee$^*$$^2$, Jinmyeong An$^{1,3}$, \\ \textbf{Yunsu Kim}$^4$,  \textbf{Jungseul Ok}$^{1,2}$, \textbf{Gary Geunbae Lee}$^{1,2}$\\
	 $^1$Graduate School of Artificial Intelligence, POSTECH, Republic of Korea \\
     $^2$Department of Computer Science and Engineering, POSTECH, Republic of Korea \\
     $^3$Mobile eXperience Business, Samsung Electronics, Republic of Korea \\
     $^4$aiXplain Inc., Los Gatos, CA, USA \\
    \texttt{\small \{solee0022, lee1jun, jungseul.ok, gblee\}@postech.ac.kr}, \texttt{\small jinmyeong.an@samsung.com}, \texttt{\small yunsu.kim@aixplain.com}
    }

\begin{document}
\maketitle
\begin{abstract}
We present DeRAGEC, a method for improving Named Entity (NE) correction in Automatic Speech Recognition (ASR) systems. By extending the Retrieval-Augmented Generative Error Correction (RAGEC) framework, DeRAGEC employs synthetic denoising rationales to filter out noisy NE candidates before correction. By leveraging phonetic similarity and augmented definitions, it refines noisy retrieved NEs using in-context learning, requiring no additional training. Experimental results on CommonVoice and STOP datasets show significant improvements in Word Error Rate (WER) and NE hit ratio, outperforming baseline ASR and RAGEC methods. Specifically, we achieved a 28\% relative reduction in WER compared to ASR without postprocessing. Our source code is publicly available at: \url{https://github.com/solee0022/deragec}.
\end{abstract}

\section{Introduction}
\def\thefootnote{*}\footnotetext{Equally contributed}\def\thefootnote{\arabic{footnote}}



Recent studies \cite{ma2023can,li2024investigating,chen2024hyporadise} have demonstrated that post-processing speech recognition transcriptions with Large Language Models (LLMs) can significantly enhance the accuracy of Automatic Speech Recognition (ASR). In particular, Generative Error Correction (GEC) has been proposed to refine ASR outputs \cite{yang2023generative, chen2024hyporadise}. Moreover, recent work has advanced GEC into a multi-modal paradigm \cite{radhakrishnan2023whispering, hularge} that jointly conditions on both the textual transcript and the corresponding audio to further improve correction quality. Under the GEC framework, an ASR model first processes the user’s speech and outputs multiple hypotheses (top-$k$) through beam search decoding. An LLM then refines these hypotheses, either by re-ranking words across different hypothesis combinations or by predicting contextually appropriate replacements. This approach leverages the extensive linguistic knowledge and powerful generation capabilities of pre-trained LLMs, which are trained on large-scale text corpora \cite{brown2020language}.

Despite its effectiveness in improving ASR accuracy, GEC faces a critical limitation: its inability to reliably introduce words absent from the initial hypotheses, particularly Named Entities (NEs) \cite{dancer, pusateri2024retrieval}. This challenge stems from inherent biases in LLMs, which, due to their exposure to large, diverse corpora, tend to favor high-frequency words and expressions \cite{gong2024contextual, gallegos2024bias}. As a result, rare or out-of-vocabulary NEs—especially those missing from the ASR hypotheses—are difficult to recover with high accuracy \cite{failing, pusateri2024retrieval}.
In response, retrieval-augmented methods \cite{dancer, failing} have recently gained traction as a way to incorporate external knowledge for more accurate NE correction. A notable example is Retrieval-Augmented Generative Error Correction (RAGEC), which augments GEC by retrieving relevant NEs from an external knowledge base and integrating them into the LLM’s input context \cite{pusateri2024retrieval}. Nonetheless, the presence of irrelevant or weakly related NEs often introduces noise, thereby undermining correction performance \cite{dancer, failing}. Current solutions typically handle this noise implicitly, relying heavily on the LLM’s internal GEC capabilities.

In this work, we propose an explicit method for denoising NE candidates, which filters out irrelevant NEs before the GEC process. Our approach combines phonetic score, augmented NE definitions, and synthetic rationales within an in-context learning (ICL) framework. By directly reducing noise in the retrieved NE candidates—without the need for additional training—we minimize the system's reliance on both ASR outputs and LLM inference. We tested our method on the CommonVoice and STOP speech datasets, achieving a 28\% relative reduction in Word Error Rate (WER).

\section{Related Works}
\subsection{Generative Error Correction in ASR} 

GEC \cite{chen2024hyporadise,yang2023generative,ma2023can} has become an effective post-processing method for ASR systems. 
However, GEC models struggle with novel or domain-specific NEs \cite{dancer,pusateri2024retrieval}. While Retrieval-Augmented GEC (RAGEC) \cite{lei2024contextualization,pusateri2024retrieval} improves NE correction by retrieving phonetically similar NE candidates, it faces challenges in determining the optimal number of retrieved NEs, leading to phonetic confusion \cite{dancer} or insufficient NEs. To address this, we propose a novel, training-free RAGEC system with an explicit denoising mechanism that generates rationales to filter and select relevant NE from the retrieved list.

\subsection{Retrieved Candidates Filtering}
RAG \cite{rag,izacard2023atlas,guu2020retrieval} improves language models (LMs) accuracy in knowledge-intensive tasks but often retrieves noisy data \cite{li2022large,yoran2023making}.
In standard RAG systems, including the RAGEC task \cite{pusateri2024retrieval,failing}, noise removal in retrieved data (e.g., NEs lists) is typically handled implicitly by training LMs to predict the correct answer (e.g., transcription) despite potentially noisy inputs. However, this approach is vulnerable to high noise ratios, lacks transparency, and heavily depends on the LMs \cite{cuconasu2024power,wu2024faithful}.
Recent studies \cite{instructrag} introduce explicit denoising by generating rationales to filter noise without additional training. However, denoising retrieved phonetic information (e.g., NE lists) in RAGEC remains underexplored.
To address this gap, we propose DeRAGEC, a training-free, rationale-driven approach that refines retrieved NE lists by explicitly denoising phonetic candidate information.

\section{DeRAGEC}
\subsection{Preliminary}  

We begin with a pre-trained ASR model that generates a 5-best hypothesis set  $H = \{h_1, h_2, h_3, h_4, h_5\}$ using beam search. From the top hypothesis $h_1$, a Named Entity Recognition (NER) model extracts NE candidates, which serve as a phonetic query $q_p$ for retrieving the top-$k$ phonetically similar NE candidates $N = \{n_1, \ldots, n_k\}$ from an external NE database.

In the \textbf{ASR baseline}, the top hypothesis is directly used as the transcription $\hat{a} = h_1$ (Eq.\ref{eq:asr}).

In the \textbf{GEC framework}, a language model $\mathcal{M}_\theta$ post-processes the hypothesis set $H$ using $T^{fs}$ number of few-shot examples $\mathcal{E}^{gec} = \{(H_i, a_i)\}$ sampled from dataset $\mathcal{D}$ (Eq. \ref{eq:gec}).
Where $\mathcal{D} = \{(H_i,a_i,N_i), 1 \leq i \leq T\}$ contains $T$ triplets of hypotheses ($H$), ground-truth transcription ($a$) and retrieved NE candidates ($N$).




The \textbf{RAGEC framework} enhances this by incorporating the retrieved NE candidates $N$ during generation, with demonstrations $\mathcal{E}^{ragec}=\{(H_i, a_i, N_i), 1 \leq i \leq T^{fs}\} \in \mathcal{D}$ (Eq. \ref{eq:ragec}) where $T^{fs}$ is number of ICL examples.

\begin{normalsize}
\begin{equation}
\textbf{ASR}: \hat{a} = h_{1}
\label{eq:asr}
\end{equation}
\end{normalsize}

\vspace{-5mm}
\begin{normalsize}
\begin{equation}
\textbf{GEC}:  \hat{a} = \mathcal{M}_{\theta}(a|H, \mathcal{E}^{gec}))
\label{eq:gec}
\end{equation}
\end{normalsize}

\vspace{-8mm}
\begin{normalsize}
\begin{equation}
\textbf{RAGEC}:  \hat{a} = \mathcal{M}_{\theta}(a|H, N, \mathcal{E}^{ragec})
\label{eq:ragec}
\end{equation}
\end{normalsize}
\begin{figure}[t]
    \centering
    \includegraphics[width=1\linewidth]{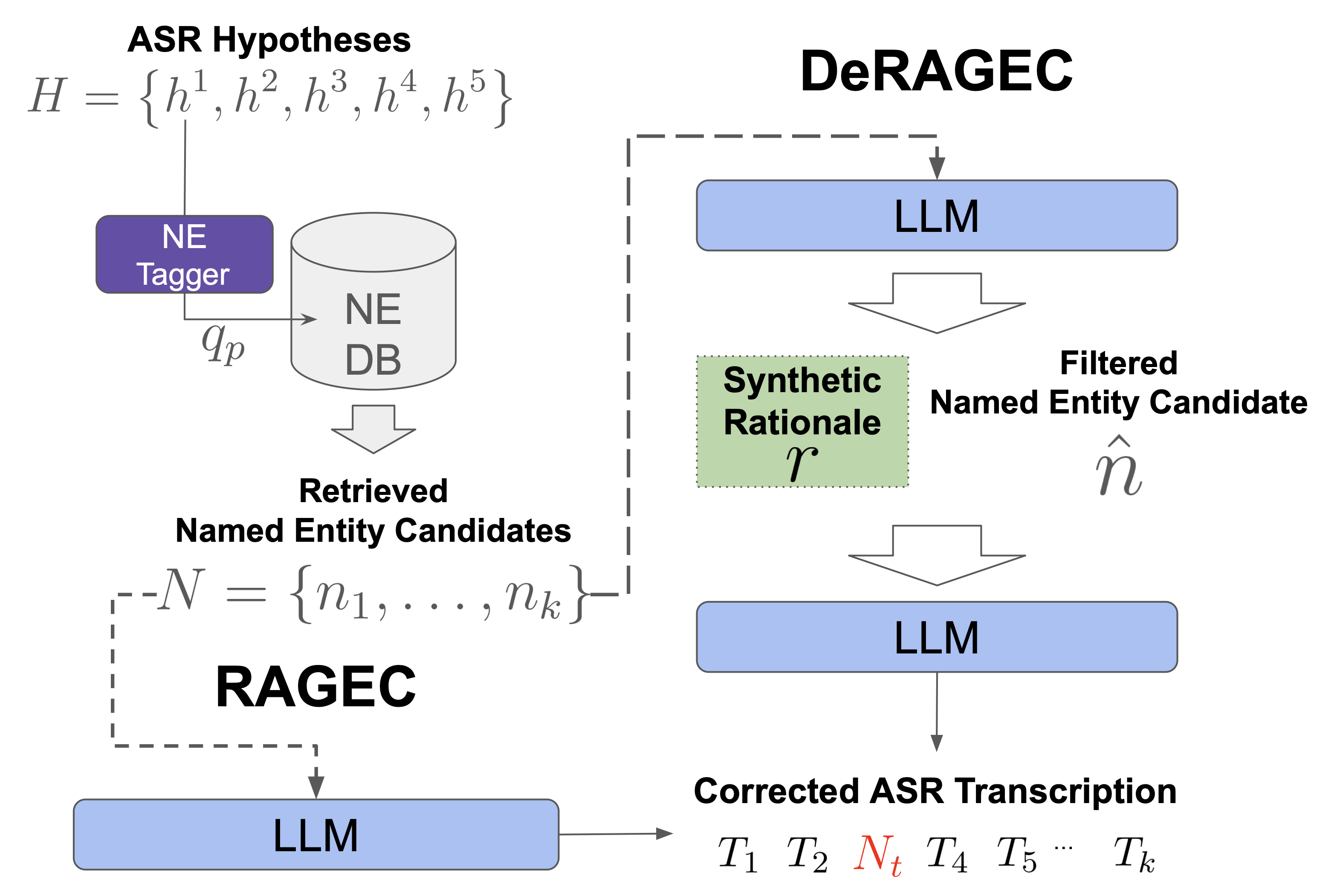}
    \caption{Comparison of RAGEC and DeRAGEC in handling retrieved NE candidates. For clarity, input features such as $PS$ and $Def$ are omitted to highlight the key differences between the two methods.}
    \label{fig:ragec}  
\end{figure}

\subsection{DeRAGEC Framework}

DeRAGEC extends RAGEC with an explicit, training-free denoising gate. For each retrieved named-entity (NE) candidate we append (i) a phonetic similarity score ($PS$) and (ii) a one-line Wikipedia definition ($Def$) to every retrieved NE candidate. 
The denoising gate prunes irrelevant candidates, returning the selected entity \(\hat{n}\) together with its rationale \(r\).  
We then pass the tuple \((H,\hat{n},r)\) to the GEC model, which outputs the final transcript \(\hat{a}\).  
Figure~\ref{fig:ragec} highlights the DeRAGEC extensions to the original RAGEC.


\paragraph{Phonetic \& Semantic Enrichment:}
For every candidate \(n_i\in N\) we attach its phonetic-similarity score  
\(PS_i = \operatorname{sim}(q_p,n_i)\) and a one-sentence entity definition \(Def_i\).
The resulting triple is serialised as:

\noindent\texttt{< $n_i$  | phonetic-score:\(PS_i\) \ | def: $Def_i$ >} 
and used for denoising gate.

\paragraph{Synthetic Rationale Generation:}
Inspired by prior work~\cite{instructrag}, we employ a separate rationale 
generation model $\mathcal{M}_r$ to synthesize denoising rationales $r^{\text{syn}}$
that guide the selection of relevant NE candidates. As illustrated in 
Algorithm~\ref{algo: rationale_gen}, for each training triplet $(h_1, a, N) \in \mathcal{D}$, along with 
$PS$ and $Def$, the model generates a MCQ-style rationale $r^{\text{syn}}$, 
inspired by~\cite{listen}, which explains which NE candidates contribute meaningfully to recovering 
the ground truth $a$. The NE list $N$ is augmented by concatenating it with the NEs extracted from $H$, 
ensuring completeness. Consequently, this allows us to effectively augment the current dataset 
$\mathcal{D} \rightarrow \mathcal{D}^+ = \{(H_i, a_i, N_i, PS_i, Def_i, r_i), 1 \leq i \leq T\}$.

\paragraph{Learning Denoising Rationales:}
DeRAGEC uses ICL to learn from these synthetic rationales without further training. Given a instance $(h_1, N, PS, Def)$, the model samples $T^{fs}$ rationale-augmented examples $\mathcal{E}^{deragec} \in D^+$ and generates:
\begin{enumerate}
    \item A filtered NE $\hat{n}$ and rationale $r$ describing the reasoning process for selecting $\hat{n}$ 
    \item The final corrected transcription $\hat{a}$, conditioned on $H$, $\hat{n}$, and $r$
\end{enumerate}

\vspace{-5.5mm}

\begin{equation}
\begin{aligned}
\textbf{DeRAGEC}: \\
\hat{n}, r &\leftarrow \mathcal{M}_{\theta}(n|(h^1,N,PS,Def), \mathcal{E}^{deragec}) \\
\hat{a} &= \mathcal{M}_{\theta}(a|H, \hat{n}, r)
\end{aligned}
\label{eq:deragec}
\end{equation}

For detailed prompts related to the rationale synthesis and GEC processes, please refer to Tables \ref{tab:module1}–\ref{tab:module3} in the Appendix \ref{sec:appendix-prompt}.

\begin{algorithm}[h!]
\caption{Synthesize Rationales}
\textbf{Input:} Phonetic Retrieval $\mathcal{R}$, Rationale Generator $\mathcal{M}_{r}$, Data $\mathcal{D}$, phonetic similarity set $PS$ and textual definitions $Def$ for NE candidates $N$.

\begin{algorithmic}[1]
\State \textcolor{blue}{/*Synthesized Rationales*/} 
\For{each $\{(H,a,N)\} \in \mathcal{D}$}
    \State Extract named entities (NE) from $H$: 
    \State $N^{hyp}=\{n^{hyp}_1,\dots,n^{hyp}_5\}$ 
    \State Concatenate $N^{hyp}$ and $N$:
    \State $N \leftarrow N^{hyp} \oplus N$
    \State Synthesize denoising rationales:
    \State $r^{syn} \leftarrow \mathcal{M}_r(h_1,a,N,PS,Def)$ 
    
\EndFor

\end{algorithmic}
\label{algo: rationale_gen}
\end{algorithm}


\section{Experimental Result}


\subsection{Experimental Setting}
For our evaluation, we use a subset of the CommonVoice (CV) dataset \cite{ardila2019common}, which includes 2,000 samples from \cite{chen2024hyporadise}, and a sub-sampled STOP test set \cite{tomasello2023stop} containing 5,000 samples. These datasets were chosen to evaluate the effectiveness of our method on both free-form speech (CV) and speech commands for spoken language understanding (STOP). We retrieved the top-10 phonetically similar NEs for RAGEC and DeRAGEC. Additionally, up to five ($T^{fs} = 5$) randomly selected samples from the CV and STOP training sets are used as ICL few-shot examples ($\mathcal{E}$).

We employ Whisper-large-v3-turbo (0.8B) \cite{radford2022whisper} with a beam search size of 5 as our \textbf{ASR} model. For \textbf{GEC}, we use Llama-3.1 (70B) \cite{meta2024llama3}, and GPT-4o-mini (\texttt{gpt-4o-mini-2024-07-18}) \cite{openai2023gpt4}. Additionally, we use o1 (\texttt{o1-2024-12-17}) \cite{jaech2024openai} for rationale synthesis.
Epitran \cite{mortensen-etal-2018-epitran} and Panphon \cite{mortensen-etal-2016-panphon} are utilized for phonemizing named entities (NEs) and computing phonetic similarity ($PS$) based on articulatory features. Specifically, phonetic similarity is measured between the tagged NE from the hypothesis (denoted as phonetic query $q_p$) and NEs retrieved from NE database ($N= \left\{n_1,\dots,n_{k}\right\}$).

A total of 3,003,462 NEs are collected from CV training set, an open-source media entity dataset \cite{van2022space}, and Wikipedia to build NE database. For NE collection and GEC process, we use GliNER-large-v2\footnote{\url{https://huggingface.co/urchade/gliner_large-v2}} model as the NE tagger.

\subsection{DeRAGEC and Baselines}

\begin{table*}[t]
\centering
\resizebox{1\textwidth}{!}{%
\begin{tabular}{c|ccccc|cccc|cccc}
\toprule
\multicolumn{1}{l|}{\multirow{2}{*}{}} & \multicolumn{5}{c|}{\multirow{2}{*}{Denoising feature}} & \multicolumn{4}{c|}{WER (\%) $\downarrow$} & \multicolumn{4}{c}{NE Hit Ratio $\uparrow$} \\
\multicolumn{1}{l|}{} & \multicolumn{5}{c|}{} & \multicolumn{2}{c}{Llama-3.1 (70B)} & \multicolumn{2}{c|}{gpt-4o-mini} & \multicolumn{2}{c}{Llama-3.1 (70B)} & \multicolumn{2}{c}{gpt-4o-mini} \\ \midrule \midrule
Model & $N$ & $MCQ$ & $PS$ & $Def$ & $Rat$ & CV & STOP & CV & STOP & CV & STOP & CV & STOP \\ \midrule
ASR only & - & - & - & - & - & 7.7 & 8.9 & 7.7 & 8.9 & 0.751 & 0.787 & 0.751 & 0.787 \\ \midrule
GEC & - & - & - & - & - & 6.8 & 7.8 & 6.9 & 7.4 & 0.782 & 0.805 & 0.784 & 0.804 \\ \midrule
\multirow{5}{*}{RAGEC (+ Denoising)} & \checkmark & - & - & - & - & 6.5 & 6.5 & 7.1 & 6.6 & 0.804 & 0.807 & 0.788 & 0.814 \\
& \checkmark & \checkmark & - & - & - & 6.5 & 6.5 & 7.1 & 6.5 & 0.804 & 0.807 & 0.788 & 0.816 \\
& \checkmark & \checkmark & \checkmark & - & - & 6.6 & \underline{6.0} & 7.1 & \underline{6.0} & 0.796 & \underline{0.828} & 0.785 & \underline{0.827} \\
& \checkmark & \checkmark & - & \checkmark & - & 6.5 & 7.2 & 7.0 & 6.9 & 0.807 & 0.697 & 0.795 & 0.752 \\
& \checkmark & \checkmark & \checkmark & \checkmark & - & 6.5 & 6.2 & 6.7 & \underline{6.0} & 0.807 & 0.815 & 0.802 & \underline{0.827} \\ \midrule
DeRAGEC (ours) & \checkmark & \checkmark & \checkmark & \checkmark & \checkmark & \textbf{6.0} & \textbf{5.9} & \textbf{6.2} & \textbf{5.8} & \textbf{0.831} & \textbf{0.838} & \textbf{0.813} & \textbf{0.842} \\ \midrule
ORACLE & \multicolumn{5}{l|}{A ground truth of NE} & 5.8 & 5.7 & 6.0 & 5.7 & 0.837 & 0.857 & 0.828 & 0.847 \\
\bottomrule
\end{tabular}%
}
\caption{Comparison of WER (\%) and NE Hit Ratio on the CV and STOP datasets across different model and denoising settings. For RAGEC, using only $N$ indicates that no additional denoising is applied, while other settings incorporate specific features to denoise NE candidates.}
\label{tab:main}
\end{table*}

To demonstrate the effect of each feature, we conducted ablation experiments on the features $MCQ$ format prompt for NE filtering, $PS$, $Def$, and rationale ($Rat$), as shown in Table \ref{tab:main} while NE candidates ($N$) is used for both RAGEC and DeRAGEC. 

Comparing the results with our baselines, including pure ASR, GEC, and RAGEC, DeRAGEC consistently achieves the best performance in both WER and NE hit ratio. 
The NE hit ratio is a metric that measures the effectiveness of the methods in correcting NEs. It is calculated by dividing the number of correctly identified NEs by the total number of NEs in the final transcription. A higher NE hit ratio indicates better performance in NE correction.

From Table \ref{tab:main}, we observe that DeRAGEC outperforms all other methods by a significant margin, achieving an average relative WER reduction of 28\% over the pure ASR baseline and 5.9\% over the best RAGEC configuration. Even when compared to the ORACLE setting, where ground truth NEs are provided to the GEC process, DeRAGEC shows a small gap in both WER and NE hit ratio, demonstrating its effectiveness even comparing with the upper bound setting.



\subsection{Effect of Denoising}
\begin{figure}[h]
    \centering
    \includegraphics[width=0.9\linewidth]{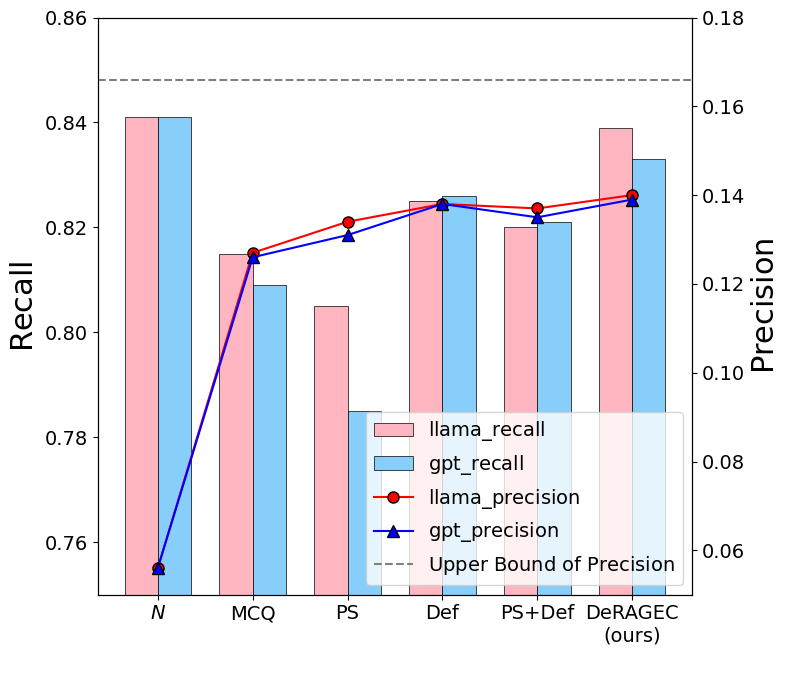}
    \caption{Recall and Precision of NE hit ratio of candidate denoising process. Different filtering methods are evaluated on CV dataset.}    \label{fig:mcq-metric}
\end{figure}

To evaluate the impact of denoising, we considered additional metrics beyond WER to assess the filtering effectiveness. Figure \ref{fig:mcq-metric} demonstrates that our method maintains the candidate hit ratio (recall) on denoising step while improving precision, suggesting that the denoising step successfully selects the correct NE and eliminates unrelated candidates. The precision upper bound is 0.166 ($1/6$), assuming the correct NE is always included in one of the selected candidates $n$, along with the 5 existing NEs in ASR hypotheses. Although our method does not reach this upper bound, it achieves the highest precision of 0.139 compared to other baselines. Additionally, with the recall upper bound set at 0.841 in $N$ only setting which dose not applied any denoising, our method maintains a recall of 0.839, demonstrating only a small gap.

\vspace{-3mm}
\section{Conclusion}
DeRAGEC is a novel, training-free approach that enhances Named Entity (NE) correction in ASR systems by filtering noisy NE candidates using phonetic similarity, augmented definitions, and synthetic rationales. Experiments on CommonVoice and STOP datasets show that DeRAGEC outperforms baseline ASR and RAGEC models, achieving a 28\% relative reduction in WER and improved NE hit ratios. These results highlight DeRAGEC's effectiveness in NE correction, even compared to an ORACLE setting, with potential for broader ASR applications.

\section*{Limitation}
Our study explored denoising retrieved named entities (NEs) using additional knowledge (phonetic and textual knowledge) and rationales within the RAGEC task, an area where denoising remains underexplored. However, our approach has some limitations. It may be necessary to investigate training methods that leverage our synthetic rationales. While this process can be costly and dependent on ASR or post-processing models, further research is needed to assess how well the model can internalize and apply the denoising instructions provided during training. In addition, it might be necessary to explore whether our approach generalizes across a wider range of ASR and post-processing models, ensuring its adaptability for broader applications. Moving forward, we aim to explore alternative denoising approaches and enhance the robustness of denoising techniques to further improve ASR post-processing effectiveness, as mentioned above.

\section*{Acknowledgments}

This work was partly supported by Institute of Information \& communications Technology Planning \& Evaluation (IITP) grant funded by the Korea government(MSIT) (No.RS-2019-II191906, Artificial Intelligence Graduate School Program(POSTECH)) (5\%) 
and was supported by the IITP(Institute of Information \& Coummunications Technology Planning \& Evaluation)-ITRC(Information Technology Research Center) grant funded by the Korea government(Ministry of Science and ICT)(IITP-2025-RS-2024-00437866) (47.5\%) and was supported by Smart HealthCare Program funded by the Korean National Police Agency(KNPA) (No. RS-2022-PT000186) (47.5\%)

\bibliography{custom}

\newpage
\appendix
\section*{Appendix}
\label{sec:appendix}

\begin{figure*}[]
    \centering
    \includegraphics[width=1\linewidth]{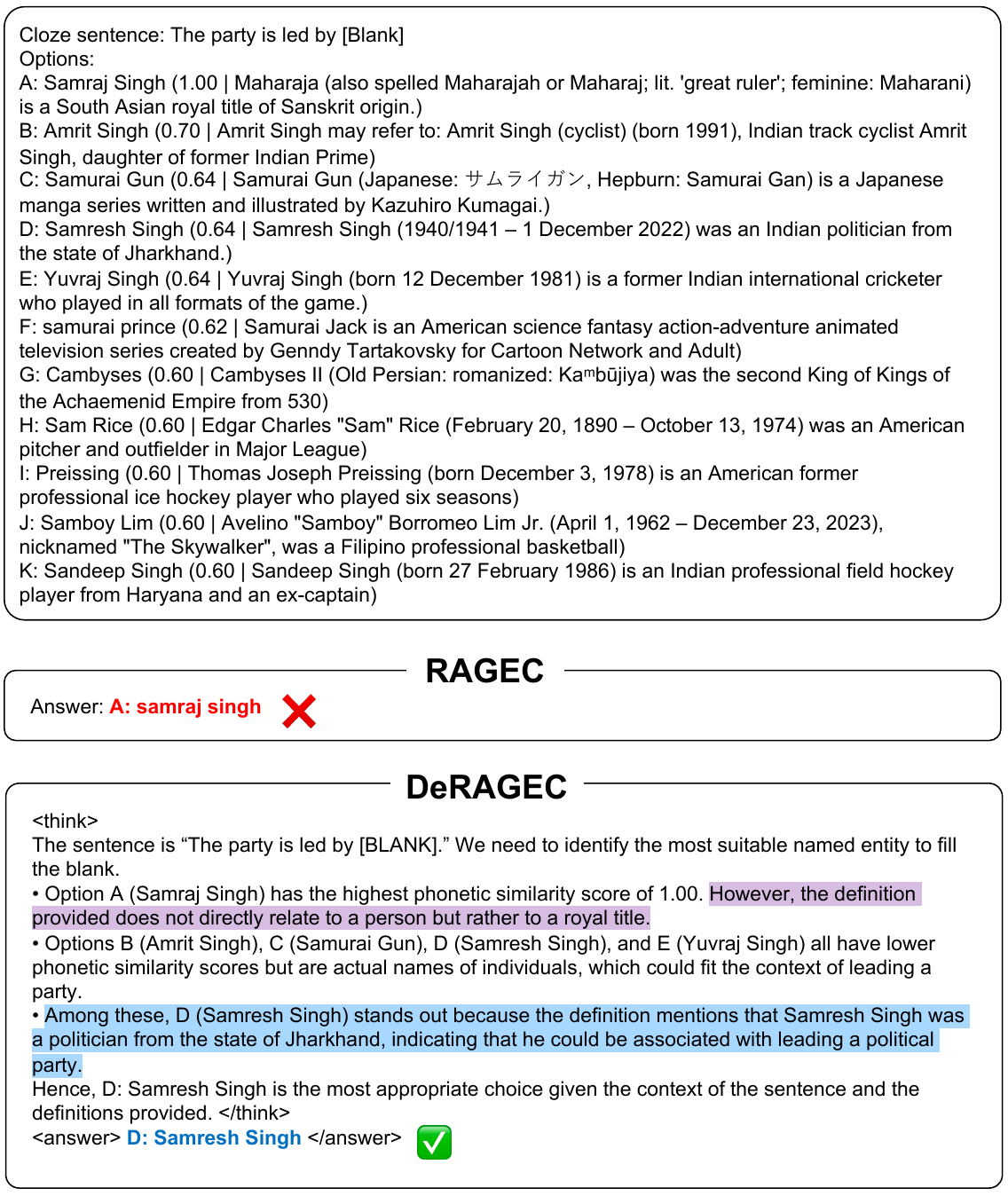}
    \caption{Case study between REGEC and DeRAGEC in CommonVoice dataset. The process of denoising irrelevant NE candidates is highlighted in pink, while the selection of the correct NE is in blue.}
    \label{fig:example1}
\end{figure*}

\begin{figure*}[]
    \centering
    \includegraphics[width=1\linewidth]{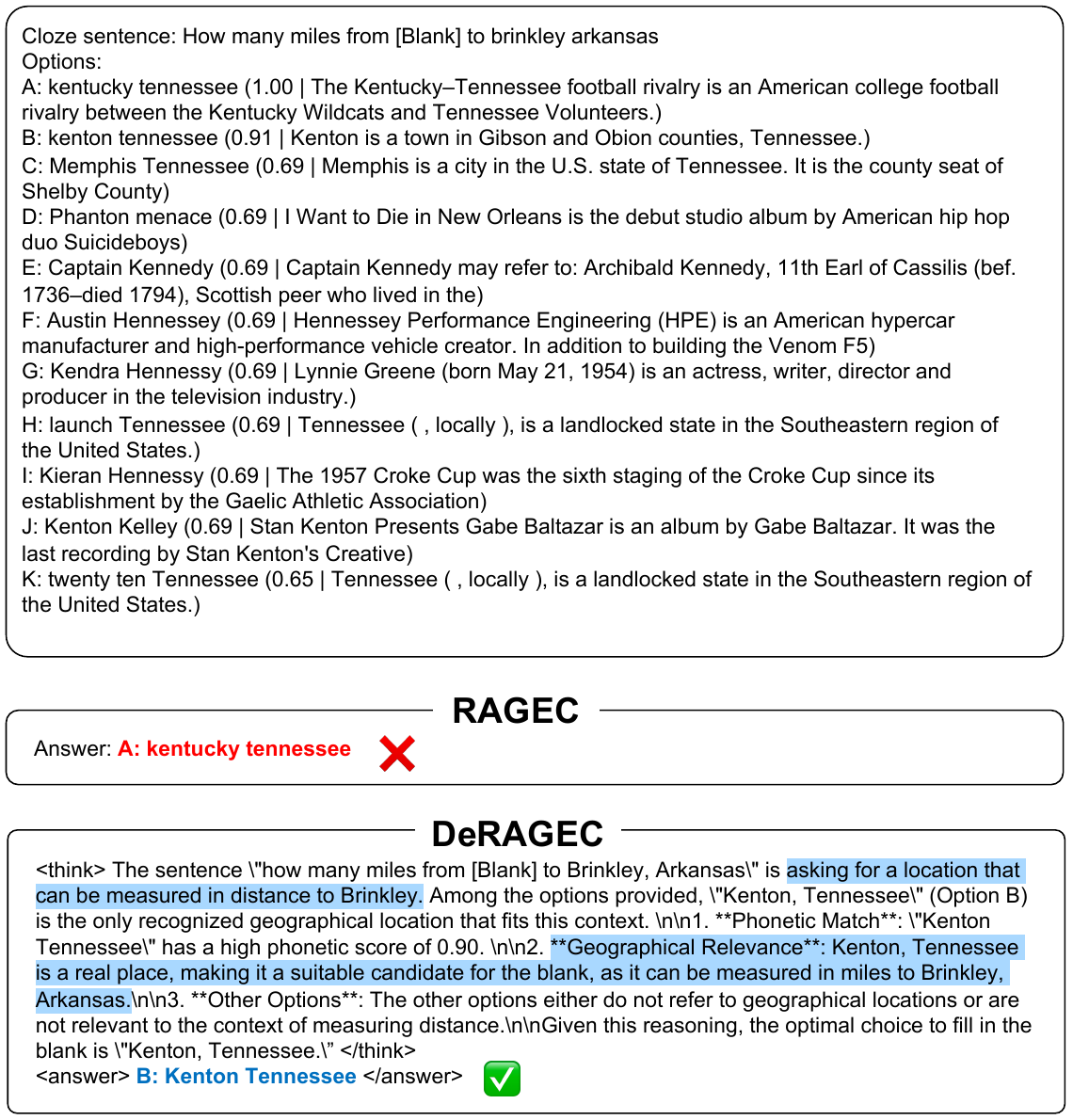}
    \caption{Case study between REGEC and DeRAGEC in STOP dataset. The process of selecting the correct NE is highlighted in blue.}
    \label{fig:example2}
\end{figure*}

\section{Effects of static threshold filtering.}
Table \ref{tab:std-cv} and \ref{tab:std-stop} present three existing methods to reduce the candidate set: reducing the K value (number of retrieved NEs, $N = \{n_1, \ldots, n_k\}$) during retrieval, setting a static distance threshold ($\theta_{d}$), and applying a statistical threshold based on the standard deviation ($STD$) of the retrieved candidates. However, the static threshold methods, including K and $\theta_{d}$, have a limitation: even correct candidates may be excluded, leading to a decrease in recall.

Instead of using a fixed constant threshold (K, $\theta_{d}$) for all cases, we also conducted an experiment using $STD$ to set a more sample-specific threshold. The statistical filtering works as follows: when the distance $STD$ of the candidates is large (indicating a skewed distribution), fewer candidates remain after filtering, specifically those within $\sigma *$$STD$  of the mean. Conversely, when the $STD$ is small (indicating a more uniform distribution), more candidates are retained. However, the statistical threshold approach did not perform well with low recall.

\begin{table*}[]
\centering
\begin{tabular}{c|c|ccc}
\hline
\multicolumn{1}{l|}{}                                                  & \multicolumn{1}{l|}{} & Recall & Precision & WER \\ \hline
\multirow{4}{*}{TopK} & 1   & 0.823 & 0.137 & 7.2 \\
                      & 5   & 0.835 & 0.084 & 6.6 \\
                      & 10  & 0.841 & 0.056 & 6.5 \\
                      & 15  & 0.843 & 0.032 & 6.7 \\ \hline
\multirow{4}{*}{\begin{tabular}[c]{@{}c@{}}PS\\ ($\theta_{d}$)\end{tabular}}   & 0.6 & 0.697 & 0.078 & 6.5 \\
                      & 0.7 & 0.457 & 0.117 & 6.6 \\
                      & 0.8 & 0.251 & 0.147 & 6.9 \\
                      & 0.9 & 0.126 & 0.163 & 7.2 \\ \hline
\multirow{4}{*}{\begin{tabular}[c]{@{}c@{}}$STD$\\ ($\sigma$)\end{tabular}} & 0.5                   & 0.830   & 0.124     & 7.0 \\
                      & 1   & 0.827 & 0.131 & 7.3 \\
                      & 1.5 & 0.822 & 0.137 & 7.4 \\
                      & 2   & 0.815 & 0.138 & 7.4 \\ \hline
\end{tabular}
\caption{Effect of static phonetic filtering in WER at CommonVoice with Llama-v3.1.}
\label{tab:std-cv}
\end{table*}

\begin{table*}[]
\centering
\begin{tabular}{c|c|ccc}
\hline
\multicolumn{1}{l|}{}                                                  & \multicolumn{1}{l|}{} & Recall & Precision & WER \\ \hline
\multirow{4}{*}{TopK} & 1   & 0.887 & 0.148 & 6.7 \\
                      & 5   & 0.895 & 0.089 & 6.7 \\
                      & 10  & 0.896 & 0.060 & 6.5 \\
                      & 15  & 0.897 & 0.043 & 6.6 \\ \hline
\multirow{4}{*}{\begin{tabular}[c]{@{}c@{}}PS\\ ($\theta_{d}$)\end{tabular}}   & 0.6 & 0.857 & 0.065 & 6.5 \\
                      & 0.7 & 0.549 & 0.105 & 6.7 \\
                      & 0.8 & 0.244 & 0.143 & 6.7 \\
                      & 0.9 & 0.035 & 0.165 & 7.2 \\ \hline
\multirow{4}{*}{\begin{tabular}[c]{@{}c@{}}$STD$\\ ($\sigma$)\end{tabular}} & 0.5                   & 0.841  & 0.107     & 6.8 \\
                      & 1   & 0.836 & 0.124 & 7.5 \\
                      & 1.5 & 0.822 & 0.134 & 7.6 \\
                      & 2   & 0.821 & 0.146 & 7.6 \\ \hline
\end{tabular}
\caption{Effect of static phonetic filtering in WER at STOP with Llama-v3.1.}
\label{tab:std-stop}
\end{table*}


\section{ORACLE comparison of different retrieval settings.}
In Table \ref{tab:oracle}, we demonstrate the denoising effect under both real retrieval situations and an upper-bound retrieval scenario, where the correct answer NE is unconditionally retrieved. For this experiment, we used the CommonVoice dataset and Llama-v3.1. In this setup, ORACLE-retr represents a scenario where the actually retrieved NE (with recall@10 of 0.841) is fed to the GEC, with noisy NEs removed. ORACLE represents a scenario where only the correct answer NE is provided to the GEC, while ORACLE+noise includes phonetically similar noisy NEs. This setup allowed us to observe the performance degradation caused by NE noise on the GEC. These results suggest that applying our methodology to various retrieval scenarios can lead to performance improvements.

\begin{table}[]
\centering
\small
\begin{tabular}{c|c|cc}
\hline
\multicolumn{1}{l|}{} & Recall                 & WER & NE hit ratio \\ \hline
RAGEC                 & \multirow{3}{*}{0.841} & 6.5 & 0.804        \\
DeRAGEC               &                        & 6.0 & 0.831        \\
ORACLE-retr           &                        & 5.8 & 0.837        \\ \hline
ORACLE                & \multirow{2}{*}{1.000} & 4.1 & 0.922        \\
ORACLE+noise          &                        & 5.4 & 0.887        \\ \hline
\end{tabular}
\caption{ORACLE comparison of different retrieval settings at CommonVoice with Llama-v3.1.}
\label{tab:oracle}
\end{table}

\section{Performance of DeRAGEC w/o MCQ.}
We attempted to concatenate all processes, reasoning, selecting the NE, and performing error correction. However, as you can see in Table \ref{tab:woMCQ}, DeRAGEC w/o MCQ show performance degradation. When DeRAGEC was performed as a single step, the generated rationale averaged 583 words, which was too long and made the process overly complicated. In contrast, the MCQ step generated an average of 211 words of rationale and performed better. Consequently, our experiments demonstrate that separating the steps for NE selection and error correction outperforms combining them into a single inference step.

\begin{table}[]
\small
\begin{tabular}{c|ccc}
\hline
\multicolumn{1}{l|}{} & RAGEC & DeRAGEC w/o MCQ & DeRAGEC \\ \hline
WER                   & 6.5   & 6.8     & \textbf{6.0}    \\
NE hit ratio          & 0.804 & 0.780   & \textbf{0.831}  \\ \hline
\end{tabular}
\caption{DeRAGEC w/o MCQ step at CommonVoice with Llama-v3.1.}
\label{tab:woMCQ}
\end{table}

\section{Case study}
Figure \ref{fig:example1} and \ref{fig:example2} compare the generated answer in the MCQ step of RAGEC and DeRAGEC. RAGEC generates only the NE, while DeRAGEC first generates a rationale to select the NE and then generates the NE itself. This study demonstrates that our model effectively denoises irrelevant NEs using $PS, Def$ and its reasoning skills, which involve identifying the relevant information, removing irrelevant NEs, and selecting the best candidate NE.

\section{Effect of Number of few-shots}
\begin{figure}[h!]
    \centering
    \includegraphics[width=1.0\linewidth]{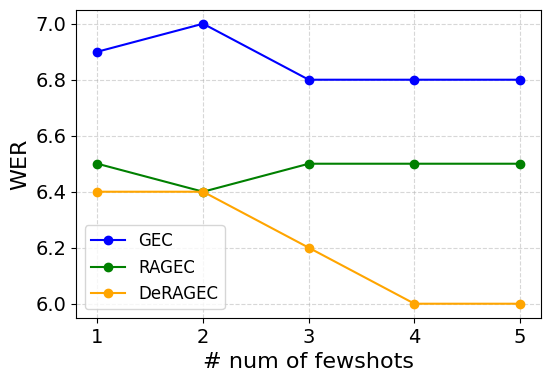}
    \caption{Effect of number of few-shot examples in WER at CommonVoice with Llama-v3.1.}
    \label{fig:fewshot}
\end{figure}

Figure \ref{fig:fewshot} shows that GEC and RAGEC has no improvement with an increasing number of fewshots. In contrast, DeRAGEC exhibits enhanced performance in generating rationales and selecting the correct NE as the number of few-shots grows. These results demonstrate that DeRAGEC effectively leverages the additional fewshots to refine its reasoning process and improve task-specific outcomes.

\section{NER performance on ASR hypothesis.} 
\begin{figure}[]
    \centering
    \includegraphics[width=1\linewidth]{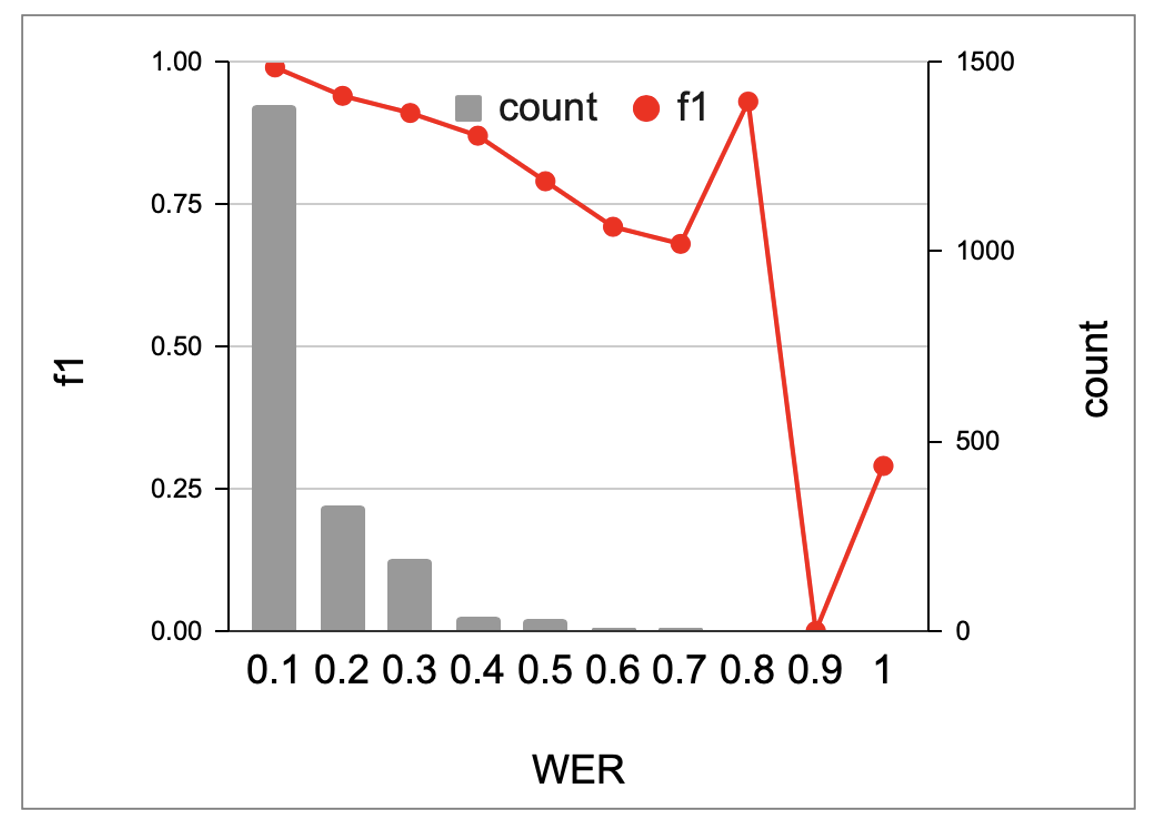}
    \caption{NER performance (F1 score) on different WER.}
    \label{fig:NER}
\end{figure}

Figure \ref{fig:NER} shows the NER performance on the ASR 1-best hypothesis, using the NER results from the Ground-Truth transcription as pseudo-labels. The NER performance decreases as the ASR hypothesis WER increases, highlighting the bottleneck in extracting phonetic queries from $h_{1}$ and retrieving potential NE candidates.

\section{Prompt templates}
\label{sec:appendix-prompt}
We modularize the framework to streamline the overall process and provide a dedicated prompt template for each module. Module 1 performs Synthetic Rationale Generation, producing intermediate explanations that serve as few-shot examples to guide the model's reasoning. Module 2 performs NE filtering by reformulating the input into a MCQ format. Module 3 conducts GEC. The corresponding prompt templates for Modules 1, 2, and 3 are detailed in Tables~\ref{tab:module1}, \ref{tab:module2-1}, \ref{tab:module2-2}, and \ref{tab:module3}, respectively.

\begin{table*}[ht]
  \centering
  \begin{tcolorbox}[
      enhanced,
      rounded corners,      
      arc=2mm,              
      boxrule=0.5pt,
      width=\linewidth,
      colframe=black,
      colback=gray!10,      
      colbacktitle=gray!100, 
      coltitle=white,       
      fonttitle=\large\bfseries,
      title=Module 1: Synthesize rationales for fewshots,
      left=2mm, right=2mm, top=2mm, bottom=2mm,
      toptitle=2mm, bottomtitle=2mm
    ]

    You are given a cloze sentence, candidate named entities, their phonetic similarity scores, and their definitions. Each named entity is provided with several options indicated by ID letters A, B, C, etc. Each option follows the format: [ID letter]: [Named-Entity] ([Phonetic similarity score] | [Definition]) \\ \\
    Your task is to identify the most appropriate Named-Entity for [BLANK].\\ Answer with the "letter: Named-Entity" corresponding to your choice A, B, C, etc. \\ 
    
    \{\textcolor{red}{input}\}\\ 
    
    Explain in detail how the input results in the answer: \{\textcolor{red}{answer}\} \\ 
    Answer should not be said at first. The rationale and answer are enclosed within <think> </think> and <answer> </answer> tags, respectively, i.e., <think> rationale here </think> <answer> answer here </answer>.
    
  \end{tcolorbox}
  \caption{Synthesize rationales for fewshots.}
  \label{tab:module1}
\end{table*}

\begin{table*}[ht]
  \centering
  \begin{tcolorbox}[
      enhanced,
      rounded corners,      
      arc=2mm,              
      boxrule=0.5pt,
      width=\linewidth,
      colframe=black,
      colback=gray!10,      
      colbacktitle=gray!100, 
      coltitle=white,       
      fonttitle=\large\bfseries,
      title=Module 2 -(1): Filtering NEs without reasoning process,
      left=2mm, right=2mm, top=2mm, bottom=2mm,
      toptitle=2mm, bottomtitle=2mm
    ]

    [Task Description] \\
    You are given a cloze sentence, candidate named entities, their phonetic similarity scores, and their definitions.
    The sentence are formatted as a cloze test, where the blanks to fill are indicated by [Blank].
    Each named entity is provided with several options indicated by ID letters A, B, C, etc. Each option follows the format: [ID letter]: [Named-Entity] ([Phonetic similarity score] | [Definition])\\ \\
    Your task is to identify the most appropriate Named-Entity for [BLANK]. Please provide the answer with [letter]: [Named-Entity] format.
    REMEMBER you should return only the answer, not return any explanation. \\
    
    I will give you few-shot examples. \\
    \{\textcolor{red}{fewshot\_examples}\}\\ 
    
    [Test Case] \\
    <input> \\
    Cloze sentence: \{\textcolor{red}{cloze\_sentence}\} \\ 
    Options: \{\textcolor{red}{options}\} \\ 
    <output> \\
    Answer:

  \end{tcolorbox}
  \caption{Filtering NEs without reasoning process.}
  \label{tab:module2-1}
\end{table*}

\begin{table*}[ht]
  \centering
  \begin{tcolorbox}[
      enhanced,
      rounded corners,      
      arc=2mm,              
      boxrule=0.5pt,
      width=\linewidth,
      colframe=black,
      colback=gray!10,      
      colbacktitle=gray!100, 
      coltitle=white,       
      fonttitle=\large\bfseries,
      title=Module 2 -(2): Filtering NEs with reasoning process,
      left=2mm, right=2mm, top=2mm, bottom=2mm,
      toptitle=2mm, bottomtitle=2mm
    ]

    [Task Description] \\
    You are given a cloze sentence, candidate named entities, their phonetic similarity scores, and their definitions.
    The sentence are formatted as a cloze test, where the blanks to fill are indicated by [Blank].
    Each named entity is provided with several options indicated by ID letters A, B, C, etc. Each option follows the format: [ID letter]: [Named-Entity] ([Phonetic similarity score] | [Definition])\\ \\
    Your task is to identify the most appropriate Named-Entity for [BLANK]. Please generate a brief explanation how the given input lead to your answer. You should not generate the answer before reasoning process. The rationale and answer are enclosed within <think> </think> and <answer> </answer> tags, respectively, i.e., <think> rationale here </think> <answer> answer here </answer>.\\
    
    I will give you few-shot examples. \\
    \{\textcolor{red}{fewshot\_examples}\}\\ 
    
    [Test Case] \\
    <input> \\
    Cloze sentence: \{\textcolor{red}{cloze\_sentence}\} \\ 
    Options: \{\textcolor{red}{options}\} \\ 
    <output>

  \end{tcolorbox}
  \caption{Filtering NEs with reasoning process.}
  \label{tab:module2-2}
\end{table*}

\begin{table*}[ht]
  \centering
  \begin{tcolorbox}[
      enhanced,
      rounded corners,      
      arc=2mm,              
      boxrule=0.5pt,
      width=\linewidth,
      colframe=black,
      colback=gray!10,      
      colbacktitle=gray!100, 
      coltitle=white,       
      fonttitle=\large\bfseries,
      title=Module 3: GEC,
      left=2mm, right=2mm, top=2mm, bottom=2mm,
      toptitle=2mm, bottomtitle=2mm
    ]

    [Task Description] \\
    You are given the 5-best hypotheses from a speech recognition system and a list of named entity candidates.
    Your task is to generate a corrected transcription using the 5-best hypotheses and named entities. \\
    
    I will give you few-shot examples. \\
    \{\textcolor{red}{fewshot\_examples}\}\\ 

    REMEMBER you should return only the corrected transcription, not return any explanation. \\

    [Test Case] \\
    <input> \\
    5-best: \{\textcolor{red}{hypotheses ($H$)}\} \\
    Named-Entities: \{\textcolor{red}{filterd\_named\_entities\_with\_generated\_rationale ($\hat{n}, r$)}\} \\ 
    <output> \\
    Corrected:

  \end{tcolorbox}
  \caption{Prompt for GEC.}
  \label{tab:module3}
\end{table*}

\end{document}